# ДИСТИЛЛЯЦИЯ НЕЙРОСЕТЕВЫХ МОДЕЛЕЙ ДЛЯ ДЕТЕКТИРОВАНИЯ И ОПИСАНИЯ КЛЮЧЕВЫХ ТОЧЕК ИЗОБРАЖЕНИЙ

А.В. Ященко, А.В. Беликов, М.В. Петерсон, А.С. Потапов

E-mail: yashenkoxciv@gmail.com

**Аннотация**

**Предмет исследования.**
Методы сопоставления и классификации изображений, а также синхронного определения местоположения и составления карты местности широко применяются на встраиваемых и мобильных устройствах. Их наиболее ресурсоемкой частью является выделение и описание ключевых точек изображений. И если классические методы выделения и описания ключевых точек могут исполняться в масштабе реального времени на мобильных устройствах, то для современных нейросетевых методов, обладающих лучшим качеством, такое использование оказывается затруднительным. Таким образом, актуально повышение быстродействия нейросетевых моделей для детектирования и описания ключевых точек. Предметом исследования является дистилляция как один из методов редукции нейросетевых моделей. Целью исследования является получение более компактной модели детектирования и описания ключевых точек, а также описание процедуры получения этой модели. **Метод.** Предложен способ сопряжения исходной и более компактной, новой модели для последующего ее обучения по выходным значениям исходной модели. Для этого новая модель обучается реконструировать выходные данные исходной модели, не используя разметку изображений. Обе сети принимают на вход одинаковые изображения. **Основные результаты.** Протестирован способ дистилляции нейронных сетей для задачи детектирования и описания ключевых точек. Предложены целевая функция и параметры обучения, обеспечивающие наилучшие результаты в рамках выполненного исследования. Введен новый набор данных для тестирования методов выделения ключевых точек и новый показатель качества выделяемых ключевых точек и соответствующих им локальных признаков. В результате обучения описанным способом новая модель, с тем же количеством параметров, показала большую точность сопоставления ключевых точек, чем исходная модель. Новая модель со значительно меньшим количеством параметров показывает точность сопоставления точек близкую к точности исходной модели. **Практическая значимость.** Предлагаемый способ позволяет получить более компактную модель для детектирования и описания ключевых точек изображений, что позволяет применять эту модель на встраиваемых и мобильных устройствах для синхронного определение местоположения и составления карты местности. Применение такой модели на серверной стороне сервиса по поиску изображений также может повысить эффективность работы сервиса.

**Ключевые слова**

глубокое обучение, детектирование ключевых точек, локальные дескрипторы изображений.

# DISTILLATION OF NEURAL NETWORK MODELS FOR DETECTION AND DESCRIPTION OF KEY POINTS OF IMAGES

**A.V. Yashchenko, A.V. Belikov, M.V. Peterson, A.S. Potapov**

## Abstract

**Subject of study.** Image matching and classification methods, as well as synchronous location and mapping, are widely used on embedded and mobile devices. Their most resource-intensive part is the detection and description of the key points of the images. And if the classical methods of detecting and describing key points can be executed in real time on mobile devices, then for modern neural network methods with the best quality, such use is difficult. Thus, it is important to increase the speed of neural network models for the detection and description of key points. The subject of research is distillation as one of the methods for reducing neural network models. The aim of the study is to obtain a more compact model of detection and description of key points, as well as a description of the procedure for obtaining this model. **Method.** A method for pairing the original and more compact, new model for its subsequent training on the output values of the original model is proposed. To do this, the new model learns to reconstruct the output of the original model without using image labels. Both networks accept identical images as input. **The main results.** A method for the distillation of neural networks for the task of detecting and describing key points was tested. The objective function and training parameters that provide the best results in the framework of the study are proposed. A new data set has been introduced for testing key point detection methods and a new quality indicator of the allocated key points and their corresponding local features. As a result of training in the described way, the new model, with the same number of parameters, showed greater accuracy in comparing key points than the original model. A new model with a significantly smaller number of parameters shows the accuracy of point matching close to the accuracy of the original model. **Practical significance.** The proposed method allows to obtain a more compact model for detecting and describing key points of images, which allows you to use this model on embedded and mobile devices for synchronous location and mapping. The use of such a model on the server side of the image search service can also increase the efficiency of the service.

## Введение

Ключевые точки – это точки на изображении, характеризующие их локальные особенности и выделяемые для последующего сопоставления. Из участков изображения, соответствующих ключевым точкам, выделяются признаковые описания, или дескрипторы, используемые далее для сопоставления изображений, классификации и других задач.

Основными критериями выбора ключевых точек являются:
- воспроизводимость, то есть возможность их повторного детектирования из других положений камеры, при изменении освещённости, масштаба и других искажениях;
- качество выделяемых признаков, а именно признаки должны позволять правильно сопоставлять точки, выделенные с разных ракурсов.

Разработано множество методов детектирования ключевых точек и выделения их признаковых описаний [1-4]. Алгоритмы на основе глубокого обучения, разработанные за последние несколько лет, превосходят традиционные алгоритмы по точности детектирования и сопоставления ключевых точек [3-5].

Применение термина «ключевая точка» справедливо и при детектировании ключевых точек лица и тела человека [6,7]. В этом случае точки связаны с определенными органами и участками тела и могут обозначать края губ, глаз и т.п. К сожалению, такой подход, а именно описание ключевых точек заранее в виде меток на произвольных изображениях, практически неприменим, из-за сложности с определением семантических свойств таких точек.

Современные алгоритмы на основе глубокого обучения для обработки изображений чаще всего основаны на сверточных нейросетевых моделях, которые способны строить представления, не уступающее, а часто и превосходящее по качеству представления, разработанные вручную. Передовые результаты, связанные с классификацией [8, 9], генерацией изображений [10], а также детектированием [11] и повторной идентификацией [12] пешеходов, получены с применением глубоких сверточных нейросетевых моделей. Такие модели могут содержать десятки слоев и сложно организованный процесс прямого распространения сигнала, однако для задач, связанных с одновременной локализацией и построением карт, справедливо ограничение для времени работы и используемой памяти таких моделей.

Применение детекторов ключевых точек для задачи одновременной локализации и построения карт подразумевает использование таких моделей на компактных устройствах, часто сильно ограниченных относительно устройств, на которых обучаются нейросетевые модели. Несмотря на то, что вывод в глубоких моделях гораздо менее требователен к вычислительным ресурсам, чем обучение, для использования нейросетевых моделей на компактных устройствах, как правило, требуется их «облегчение».

Существует несколько подходов для «облегчения», упрощения или прореживания (pruning) моделей глубокого обучения [13, 14]. Большинство из них основано на предположении, что нейросетевые модели избыточно параметризированы и что удаление искусственных нейронов, не вносящих большой вклад в результат целевой метрики, можно осуществить без значительной потери качества, а иногда даже получить улучшение целевой метрики, например точности распознавании. Этот эффект можно объяснить тем, что прореживание равносильно регуляризации сети. Как правило, алгоритмы такого рода прореживания включают операции «физического» удаления отдельных элементов сети — нейронов в полносвязной сети или фильтров в сверточных архитектурах.

Другие подходы могут быть основаны на сжатии фильтров сверточной сети в частотной области [15] или обучении новой, более компактной модели, используя подход на основе дистилляции исходной модели, которую необходимо упростить [16]. Последний подход применяется в нашем исследовании. Кроме того, анализ моделей глубокого обучения остается особым ремеслом и, как показано в [17], может осуществляться с помощью других глубоких моделей.

Цель данной работы состоит в разработке и исследовании метода сопряжения исходной и новой моделей для дальнейшего обучения новой модели, а также получения целевой функции и набора данных для обучения и оценки качества работы модели.

## Описание исходной модели

В данной работе представлен алгоритм, использующий базовую модель (учитель) для обучения новой модели с меньшим числом параметров (ученик), а также результаты обучения новой модели. В качестве модели учителя выбрана модель, представленная в [3] – SuperPoint, которая представляет собой полносверточную сеть, принимающую полноразмерное изображение и порождающую два типа выходных данных: тензор (многомерный массив) для последующего детектирования ключевых точек и тензор, содержащий дескрипторы для каждой области 8×8 пикселей. Таким образом, для задачи детектирования и описания ключевых точек большая часть сети объединена, однако для порождения ключевых точек и дескрипторов, сеть-учитель разделяется на два «рукава», содержащие данные для извлечения ключевых точек и дескрипторов. Дескрипторы для точек получаются с помощью билинейной интерполяции тензора дескрипторов в координатах полученных ключевых точек.

Процесс вывода в модели SuperPoint проиллюстрирован на рисунке 1. Этот процесс включает несколько этапов. Сначала, входное изображение подается кодировщику (а). Используя представление (блок z), порожденное кодировщиком, вычисляются выходные карты признаков (б), используемые для детектирования ключевых точек (блок КТ) и выделения дескрипторов (блок ДК). Выходные карты признаков КТ и ДК обрабатываются дополнительно как было описано выше.

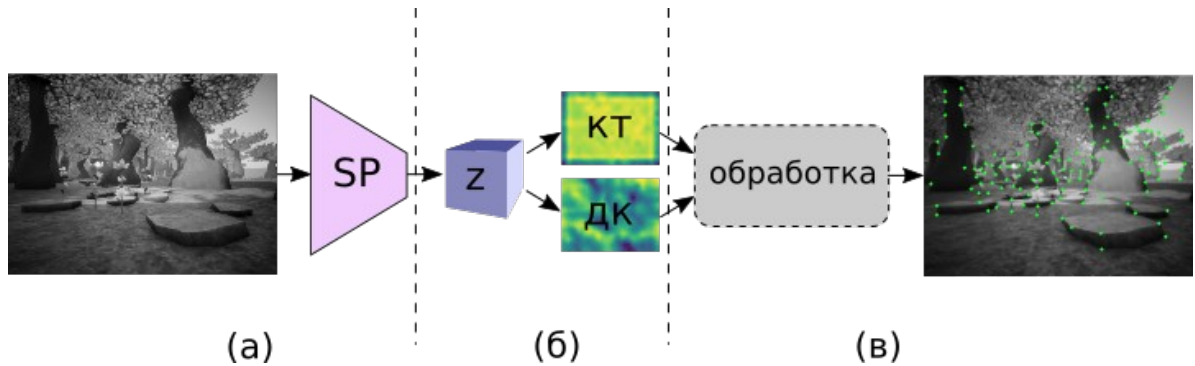

**Рисунок. 1** Процесс вывода в модели SuperPoint

Кодировщик, производящий общее представление (z), позволяет последовательно снижать размерность входного изображения с помощью блоков, включающих сверточные слои, операции субдискретизации и применение нелинейной активационной функции. Кодировщик разработан таким образом, что каждый элемент карты признаков (z) содержит "ячейку" размером 8×8, а детектирование ключевых точек проводится с помощью нормированной экспоненциальной функции (softmax) к каждой "ячейке". Далее после удаления канала, указывающего на отсутствие ключевой точки, карта признаков преобразуется к форме входного изображения.

Обучение модели предполагает оптимизацию сложной функции ошибки, которая состоит из двух компонент: 1) ошибки детектирования ключевых точек – $L_p$; 2) ошибки соответствия дескрипторов – $L_d$. Для каждого входного изображения формируется пара – второе изображение, полученное после применения гомографического искажения к первому изображению, параметризованного случайно. Таким образом, появляется пара изображений и соответствие ключевых точек для этих изображений. Формируется часть функции ошибки $L_p = L_p(X, Y) + L_p(X', Y')$, где $L_p$ – кроссэнтропия, минимизация которой означает, что сеть справляется с детектированием точек на исходном и искаженном изображении.

Процесс создания функция ошибки для дескрипторов состоит из трёх этапов:
- ошибка для ожидаемых сопоставлений $d_{прав}(desc1, desc2) = 1 - cos(desc1[ids], desc2)$, где ids - ожидаемые индексы сопоставлений дескрипторов
- ошибка для неправильно сопоставленных дескрипторов $d_{ошиб}(desc1, desc2) = cos(desc1[ids], desc2)$, где ids - индексы сопоставлений дескрипторов, не совпадающие с ожидаемыми
- ошибка для случайных сопоставлений $drand(desc1[i], desc2[j]) = cos(desc1[i], desc2[j])$ если $cos(desc1[i], desc2[j]) > 0.2$, иначе 0, где i,j - случайные индексы, cos - косинус угла между дескрипторами

Итоговая функция ошибки $L_d = d_{прав} + d_{ошиб} + d_{rand.}$

Ожидаемые сопоставления вычисляются путём проецирования точек из плоскости изображения X в плоскость изображения X' и вычисления ближайших соседей.

Описанный выше подход используется для обучения исходной модели — учителя. При этом исходный код недоступен, и воспроизвести результаты, используя исходную кодовую базу, продемонстрированные в [3] невозможно. Более того, обучение модели с меньшим количеством параметров «с нуля» может привести к менее стабильному процессу обучения и потребует корректировки значений гиперпараметров. Для решения указанных проблем предлагается метод на основе дистилляции нейронных сетей описанный далее.

**Описание предлагаемого метода и эксперименты**

Основная цель экспериментов заключается в получении более компактной модели, позволяющей эффективно детектировать и сопоставлять ключевые точки. В качестве базовой модели, подлежащей "сжатию", используется описанный ранее SuperPoint. Для получения более компактной модели используется алгоритм дистилляции или разновидность обучения с учителем (rich-supervision), суть которого заключается в попытке обучить новую, обычно более компактную, модель по выходам базовой сети. Получается, что для входного изображения базовая модель порождает некоторое представление, используемое в функции ошибки для обучения новой модели.

Функция ошибки для обучения или дистилляции новой модели, как правило, включает некоторую меру расстояний, минимизируя которую, обучается новая модель. Один из экспериментов подразумевает воспроизведение базовой модели с новым алгоритмом обучения. Однако основная цель заключается в получении меньшей модели со сравнимым критерием качества относительно базовой модели.

Для воспроизведения модели предлагается использовать такую же архитектуру сети как для базовой модели. Обозначим базовую модель как B, а ее выходы для ключевых точек и дескрипторов – кт$_в$ и дк$_в$ соответственно. Новая модель, T, воспроизводит работу B, но параметры T представлены случайными числами, близкими к нулю. Подход к сопряжение моделей T и B проиллюстрировано на рисунке 2.

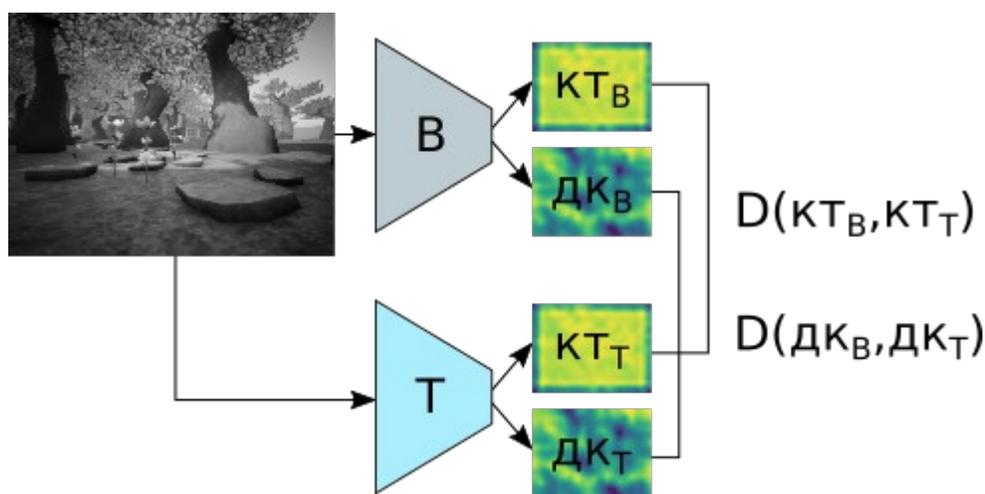

**Рисунок. 2** Сопряжение базовой и новой модели

Для обучения модели T предлагается минимизировать среднеквадратичное отклонение выходных тензоров кт$_в$ и кт$_т$. Для обучения воспроизведения дескрипторов также предлагается минимизировать среднеквадратичное отклонение выходных тензоров дк$_в$ и дк$_т$. Таким образом, функция ошибки для обучения модели T имеет вид L = D(кт$_в$, кт$_т$) + D(дк$_в$, дк$_т$). Функция ошибки является дифференцируемой и может использоваться вместе с алгоритмом градиентного спуска для оптимизации параметров модели T.

Для обучения новой модели использовались синтетические изображения, полученные с помощью средств моделирования AirSim [18].

Точность сопоставления для базовой модели составляет 0.74562, а для новой модели – 0.75151. Примечательно, что модель с идентичной архитектурой и описанным алгоритмом обучения позволяет получить несколько лучший результат метрики полноты. Однако основная цель экспериментов заключается в получении более компактной модели T. Для этого была обучена модель с вдвое меньшим количеством фильтров, чем базовая модель, метрика полноты для этой модели составила 0.74102. В результате уменьшения количества фильтров осталось только 23 % от общего количества параметров исходной модели.

Так как базовая модель является полносверточной и производит карты признаков, содержащие "ячейки" исходного изображения, то минимизация среднеквадратичного отклонения по каждому элементу таких карт признаков может приводить к "размытию" карт признаков, производимых моделью T. Предлагается минимизировать среднеквадратичное отклонение пространственных градиентов (применяется фильтр Собеля) для кт$_в$ и кт$_т$. Тогда новая функция ошибки: L = D(кт$_в$, кт$_т$) + D(дк$_в$, дк$_т$) + D(G(кт$_в$), G(кт$_т$)). При дополнении функции ошибки, модель T, будучи идентичной B, дает наилучший результат метрики полноты – 0.75842.

## Генерация базы изображений и тестирование метода

Для обучения модели на изображениях с различными условиями освещения в работе использовался набор виртуальных сцен на базе программного обеспечения Unreal Engine, а также библиотека AirSim [18]. Данная библиотека предоставляет инструментарий для получения изображений виртуальных сцен и соответствующих карт глубины с задаваемыми внутренними и внешними параметрами камеры. Эти данные использовались нами в качестве опорных для оценки точности сопоставления локальных признаков, полученных на выходе исследуемых моделей. Пример изображений и карт глубины, полученных из базы AirSim представлен на рисунке 3.

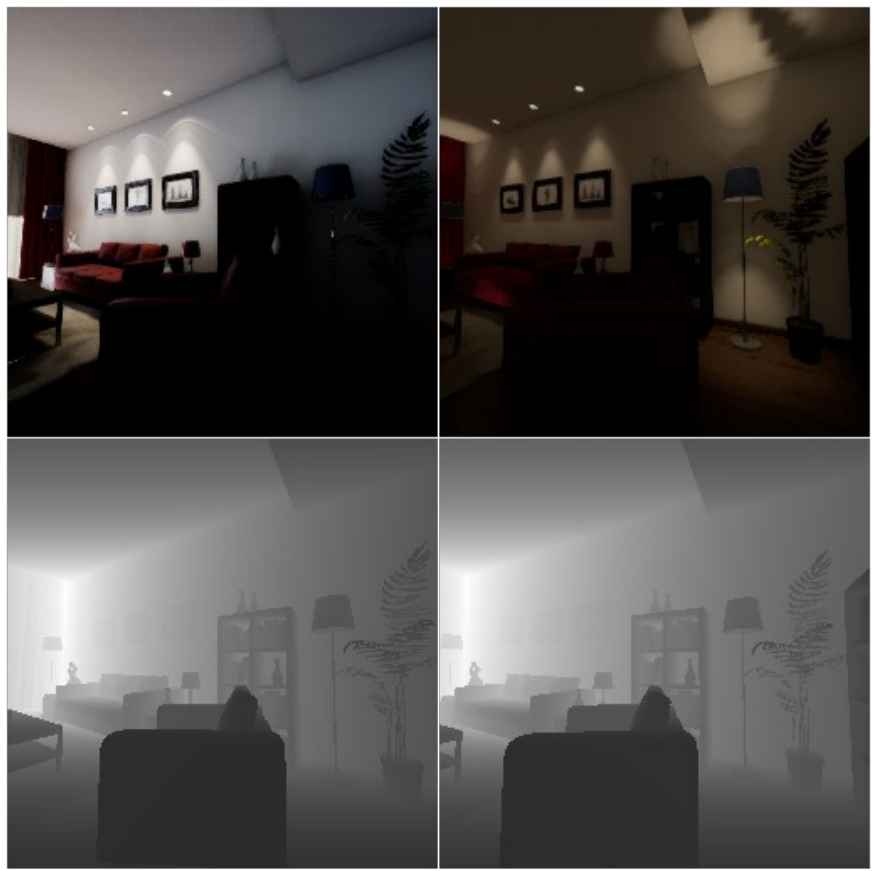

**Рисунок. 3**. Визуализированная в AirSim сцена с картами глубины.

Для каждого изображения $I_i$ в базе хранится соответствующая ему карта глубины $D_i$, внутренние параметры камеры $K_i$, внешние параметры: матрица поворота $R_i$, задающая ориентацию камеры в мировых координатах, а также вектор смещения $t_i$. Траектории камеры для последовательностей при разных условиях освещения совпадают. Матрица внутренних параметров камеры [19] фиксирована и определена в AirSim как: K = [[W/2, 0, W / 2], [0, W/2, H / 2], [0, 0, 1]], где W, H – ширина и высота изображения.

Стоит отметить, что используемая версия AirSim v1.2.0 по умолчанию вместо нормальных расстояний от точек сцены по оси Z в системе отсчета камеры возвращает расстояния до проекционного центра камеры. Поэтому для получения Z координат наблюдаемых точек сцены необходима дополнительная конвертация. Тогда, исходя из известных параметров камеры, мы можем восстановить трехмерные координаты ключевых точек изображения $I_i$ в системе отсчета i-й камеры, перенести их в систему отсчета i+1-й камеры, спроецировать эти точки на изображение $I_{i+1}$, оценить расстояния до соответствующих сопоставленных точек, и в итоге рассчитать точность сопоставления для данной пары изображений.

Для обучения модели, совмещающей выделение ключевых точек и вычисление дескрипторов желательно иметь единственный критерий качества модели. Дело в том, что при обучении модели существует баланс между компонентами целевой функции. Предлагается использовать гармоническое среднее с использованием точности дескрипторов и полноте детектора. Полнота детектора соответствует воспроизводимости ключевых точке.

Гармоническое среднее даёт большие значения для множества незначительно отличающихся величин, чем для множества с большой разницей, в отличие от арифметического среднего. Значения гармонического среднего для ключевых показателей представлены в таблице 1.

В работе [5] приводится точность и полнота для оценки качества дескрипторов, и воспроизводимость для оценки качества детектора ключевых точке. Полнота определяется как: recall = количество правильных сопоставлений / количество соответствий.

Эта величина обладает следующими особенностями. При попиксельном вычислении числа соответствий в знаменателе всегда будут большие числа, пропорциональные площади изображения. Использование всех участков изображения для сравнения дескрипторов не отражает качество работы системы, так как в практических приложениях, как правило, дескрипторы выделяются только с релевантных точек.

Метрика для оценки качества сопоставления – это отношение количества правильно сопоставленных пар локальных признаков, к общему количеству сопоставленных пар. Данная величина аналогична точности (precision), применяемой при оценке результата информационного поиска. Здесь мы также будем использовать термин «точность» для применяемой нами метрики. В таком определении точность будет зависеть от выбора исходного изображения, так, например, если на изображении $I_1$ выделена одна точка (и она правильно сопоставлена) а на $I_2$ – сто, то точность будет 1.0 для сопоставления из $I_1 \rightarrow I_2$ и 0.01 для $I_2 \rightarrow I_1$.

Precision = TP / (TP+FP), где TP – число правильно сопоставленных по дескрипторам точек, FP – неправильно сопоставленных. Эту зависимость можно устранить, взяв среднее значение точности от прямого и обратного сопоставлений.

Также важным показателем качества является воспроизводимость, равная доле точек одного изображения, детектируемых на другом. Эта метрика идентична точности при использовании идеального генератора дескрипторов, т.е. когда все точки сопоставляются правильно. Повторяемость для точек соответствует полноте (recall) в оценке информационного поиска. Для повторяемости также имеет смысл считать среднюю величину для сопоставлений $I_1 \rightarrow I_2$ и $I_2 \rightarrow I_1$.

Для информационного поиска Recall = TP / (TP + FN), для ключевых точек обозначим Recall(points($I_1$), points($I_2$)) = (points($I_1$) ∩ points($I_2$)) / points($I_2$), то есть доля точек на $I_2$ воспроизведённая на $I_1$. Имея точность и полноту можно посчитать F1 = 2 * (precesion(d, p) * recall(p)) / (precesion(d, p) + recall(p)).

Пара локальных признаков считается верно сопоставленной, если ошибка репроекции ключевых точек с первого изображения в паре на второе изображение не превышает заданного порога. В нашем случае был установлен порог в три пикселя.

База [20] содержит два типа наборов изображений: изображения с яркостной изменчивостью и изображения с ракурсной изменчивостью. Результаты тестирования представлены в таблице 2, а для базы AirSim [18] в таблице 3 и 4. Таким образом, видно, что качество выделения ключевых точек изменилось незначительно.

**Заключение**

Рассмотрено применение метода дистилляции к моделям на основе глубокого обучения для задачи детектирования и описания ключевых точек. Предлагается алгоритм обучения новой модели детектирования и описания ключевых точек на основе существующей. В предложенном алгоритме, новая модель обучается воспроизводить (дистиллировать) выходные данные базовой модели. Для этого минимизируется сложная функция ошибки, включающая меру расстояния для тензоров ключевых точек и дескрипторов, а также регуляризация в виде пространственного градиента для карт признаков ключевых точек. К недостатку текущего алгоритма можно отнести зависимость от качества синтетических данных. Дальнейшим направлением исследований мы ставим переход к обучению без учителя, что может устранить данный недостаток.

В результате экспериментов не удалось получить идентичную модель – значения функции ошибки всегда больше нуля. Однако по целевой метрике – точности сопоставления ключевых точек, – новая модель превосходит базовую. Точность сопоставления более компактной новой модели отличается незначительно от точности исходной модели.

## Литература

| точность | полнота | гармоническое среднее(F1) | арифметическое среднее |
|---|---|---|---|
| 0.61 | 0.55 | 0.58 | 0.58 |
| 0.65 | 0.51 | 0.57 | 0.58 |
| 0.16 | 1 | 0.28 | 0.58 |

Таблица 1. Сравнение интегральных критериев F1 и арифметического среднего

| Модель | Воспроизводимость детектирования, отн. ед. | | Точность сопоставления, отн. ед. | |
|---|---|---|---|---|
| | Яркостная изменчивость | Ракурсная изменчивость | Яркостная изменчивость | Ракурсная изменчивость |
| Оригинальная модель SuperPoint | 0.610781 | 0.536998 | 0,7909 | 0,7139 |
| Редуцированная модель | 0.565761 | 0.490540 | 0,7953 | 0,6802 |

Таблица 2. Сравнение редуцированной и оригинальной модели в условиях яркостной или ракурсной изменчивости

| | точность дескрипторов | повторяемость точек | гармоническое среднее |
|---|---|---|---|
| Оригинальная модель SuperPoint | 0.7515 | 0.4482 | 0.5615 |
| Редуцированная модель | 0.7192 | 0.4492 | 0.5530 |

Таблица 3. Результаты тестирования модели на AirSim Village

| | точность дескрипторов | повторяемость точек | гармоническое среднее |
|---|---|---|---|
| Оригинальная модель SuperPoint | 0.8829 | 0.5500 | 0.6778 |
| Редуцированная модель | 0.8563 | 0.54359 | 0.6650 |

Таблица 4. Результаты тестирования модели на AirSim Fantasy Village